\documentclass[sigconf,nonacm,9pt]{acmart}

\usepackage{booktabs}
\usepackage{enumitem}
\usepackage{xspace}
\usepackage{balance}
\usepackage{tikz}
\usetikzlibrary{arrows.meta, positioning, calc, fit, backgrounds, shapes.geometric}
\usepackage[subtle]{savetrees}

\newif\ifmarkrev \markrevfalse 
\newcommand{\rev}[1]{\ifmarkrev{\color{blue}#1}\else#1\fi}

\newcommand{\evolve}{altk-evolve\xspace}

\setcopyright{none}
\renewcommand\footnotetextcopyrightpermission[1]{}
\begin{document}

\title{Memory as Middleware for Self-Improving AI Agents}

\author{K. R. Jayaram, Vatche Isahagian, Vinod Muthusamy, Gegi Thomas,
        Punleuk Oum, Gaodan Fang, Ashwath Vaithinathan Aravindan}
\affiliation[obeypunctuation=true]{%
  \institution{IBM Research}, \country{USA}}

\renewcommand{\shortauthors}{Jayaram et al.}

\begin{abstract}
AI agents are stateless \rev{across sessions} by default and \rev{therefore} operationally amnesic: each session begins with little durable knowledge of prior failures, repairs, preferences, or successful strategies. As a result, agents repeat the same mistakes and discard hard-won experience. The dominant fix is \emph{bespoke memory}---retrieval, persistence, and learning logic hand-wired into one agent and bound to one storage engine. This creates a fragmented landscape where memory cannot be swapped, shared, isolated, or reasoned about independently of the agent that owns it. We argue that this is a middleware problem: agent memory deserves a first-class, pluggable layer, just as data access, messaging, and persistence each became middleware concerns.

We develop this vision through six systems challenges: two-sided pluggability, host-native interposition, multi-tenant isolation, write-path consistency, federated sharing with provenance, and lifecycle governance. We present \evolve, a reference implementation of memory middleware for self-improving agents, and use it to motivate a broader research agenda for future memory middleware.
\end{abstract}

\ccsdesc[500]{Computing methodologies~Intelligent agents}
\ccsdesc[300]{Computing methodologies~Multi-agent systems}
\ccsdesc[300]{Information systems~Information storage systems}
\ccsdesc[100]{Security and privacy~Access control}

\keywords{agent memory, memory middleware, LLM agents, self-improving agents,
  pluggable storage, multi-tenant isolation, write-path consistency,
  provenance, memory lifecycle governance, Model Context Protocol}

\maketitle

\makeatletter
\fancyhead[LE]{\@headfootfont IBM Technical Report}
\fancyhead[RO]{\@headfootfont IBM Technical Report}
\makeatother

\noindent\emph{Status of this document.} This is a preliminary IBM Technical
Report. The content is under active revision and this document will be
updated; please check for a later version before citing it.

\section{Introduction}

Large-language-model (LLM) agents are, by construction, amnesiac. A coding assistant that spends an hour discovering a required build flag, a missing sandbox tool, or an unusual pagination scheme rediscovers all of it from scratch in the next session: the context window is working memory, not durable memory. As agents move from one-shot chat interfaces to long-running software actors that plan over many steps, invoke tools, and cooperate with other agents~\cite{yao2023react,webarena2024,appworld2024,sweagent2024}, this amnesia stops being a curiosity and becomes a reliability ceiling. Such agents face the same operational phenomena as traditional distributed software---repeated failures, partial repairs, environmental quirks, and deployment-specific conventions---which software organizations carry forward through logs, runbooks, and patches. Agent systems increasingly expose rich execution traces, but rarely provide a reusable infrastructure layer that turns those traces into durable behavioral memory.

The prevailing response is to add memory \emph{inside} the agent---a vector store, a retrieval prompt, and an ad hoc ``save what worked'' step wired into the control loop. This works, and prior work shows that agents benefit from such memories, reflections, skill libraries, and trajectory-derived guidance~\cite{shinn2023reflexion,park2023generativeagents,wang2023voyager,fang2026trajectory}. But the resulting memory is \textbf{bespoke}: coupled to one agent's code, bound to one storage engine, scoped to one user, and invisible to every other agent, so each team rebuilds the same plumbing and none of it composes (Section~\ref{sec:problem}).

\rev{The systems community has been here before.} Data access, message passing, transactions, caching, and persistence were each, at some point, hand-wired into applications---until the systems community extracted them into middleware: reusable layers with uniform contracts, pluggable providers, and well-understood guarantees. We argue that \textbf{agent memory is the next such layer}, and that the middleware community is the right one to build it. A memory store persists information; memory middleware defines how experience is captured, consolidated, isolated, retrieved, shared, edited, forgotten, and governed across agents. \rev{Such external memory complements in-weight learning, RLHF, and guardrails rather than competing with them: it captures deployment-, project-, and organization-specific operational knowledge that cannot practically be trained into model weights per user or per tenant. Our scope is software agents that act through tools over long-lived operational contexts---coding, workflow, and enterprise agents; embodied agents, whose memory is coupled to perception and control, raise different constraints that we do not address.}

{\bf Our Big Idea:} Agent memory should be a pluggable middleware substrate that sits \emph{between} agents and storage, exposes one contract to many hosts (agent runtimes such as Claude Code or Codex) and many backends, and owns the cross-cutting concerns---isolation, consistency, sharing, provenance, user control, and lifecycle governance---that no individual agent should re-implement. We develop this vision as six systems challenges, framed around three design goals---\emph{accuracy}, \emph{cost}, and \emph{controls}, of which cost and controls are what make memory a systems problem rather than only a modeling one---and detailed in Section~\ref{sec:design-space}. As a first step toward this vision we present \evolve\footnote{\evolve is open source and available at \url{https://github.com/AgentToolkit/altk-evolve}.}~\cite{altkevolveRepo}, a reference implementation of memory middleware for self-improving agents that captures agent trajectories, distills them into reusable guidelines, and serves them through pluggable storage and a standard tool protocol. This paper makes four contributions:
\begin{enumerate}[leftmargin=*]
    \item We identify \emph{agent memory middleware} as an emerging systems problem: the need to decouple memory generation, storage, retrieval, consolidation, and sharing from individual agent implementations.
    \item We articulate a middleware design space for self-improving agents, organized around two-sided pluggability, host-native interposition, multi-tenant isolation, write-path consistency, federated sharing with provenance, and lifecycle governance.
    \item We show that \evolve realizes an initial subset of this design space, using trajectory capture, guideline extraction, pluggable storage, standard tool interfaces, namespaces, visibility controls, provenance metadata, and conflict-aware updates.
    \item We provide a preliminary evaluation showing that this middleware factoring improves agent reliability on AppWorld~\cite{appworld2024}, a public multi-step benchmark, and use the implementation's current limits to motivate a research agenda spanning consistency, memory admission, deployment-aware packaging, eviction, sharing, provenance, user control, and portability.
\end{enumerate}

\section{Why Agent Memory is a Middleware Problem}
\label{sec:problem}

The phrase \emph{agent memory} is overloaded: it can mean raw transcripts, user preferences, vectorized documents, episodic summaries, semantic facts, tool traces, agent manifests, or learned behavioral rules. This ambiguity obscures a more important distinction---between memory as \emph{stored data} and memory as \emph{infrastructure}---that the rest of this section develops.

\subsection{Bespoke Memory $\neq$ Scalable Infrastructure}

Current agent systems treat memory as part of the agent's application logic: a developer decides which traces to save, how to summarize and persist them, what to retrieve and inject, and how conflicts are resolved. Such mechanisms demonstrably help---agents reflect on failures, accumulate skills, and extract trajectory-derived guidance~\cite{shinn2023reflexion,park2023generativeagents,wang2023voyager,fang2026trajectory}, and coding-agent ecosystems expose persistent instruction surfaces such as project manifests, \texttt{AGENTS.md} files, and \texttt{CLAUDE.md}-style guidance~\cite{chatlatanagulchai2025manifests,codexAgentsMd,agentsMd,antigravityKnowledge}. But they are local conventions: the write path, read path, storage format, retrieval policy, and governance model are all coupled to one agent, harness, or store---easy to prototype, hard to operate as shared infrastructure.

This coupling creates several problems. First, memory is not portable: moving an agent from one framework or backend to another requires rewriting memory logic or translating ad hoc files and prompts. Second, memory is not composable: multiple agents cannot reliably share useful experience without also sharing implementation details. This is already visible in coding workflows, where a developer may use Claude Code, Codex, Antigravity, and other coding agents over the same repositories, while project preferences and learned repair strategies remain scattered across tool-specific memories and instruction files. Third, memory is not semantically stable: a retrieved item may be a fact, preference, warning, repair, policy, or obsolete artifact, but the surrounding system may treat all of them as text chunks. Fourth, memory is not governable in either the enterprise or personal sense: organizations need visibility, policy, and audit controls, while individual users need the ability to inspect, edit, delete, approve, and scope what an agent remembers about them or their projects~\cite{openaiMemoryFAQ}.

\subsection{Stores Are Not Middleware}

The growing availability of memory stores is important but insufficient. Local files, transcript databases, vector indexes, managed memory services, and long-term personalization stores provide persistence and retrieval~\cite{openaiMemoryFAQ,awsAgentCoreMemory,googleMemoryBank,cloudflareAgentMemory}. Middleware must provide the semantics above them. It must define the operations by which trajectories become memories, memories become context, and context changes future behavior. It must also define the control plane through which memories are isolated, shared, edited, expired, audited, and overridden.

The distinction matters because agent memory has multiple writers. Memories may be written by the agent itself after observing a trajectory; by a user who records preferences, corrections, or project-specific conventions; by another agent that discovers a useful repair; or by an organization that publishes rules, policies, security constraints, and guardrails. A memory store can persist all of these items, but it does not by itself define precedence, provenance, conflict handling, approval, or visibility. Middleware should specify how these heterogeneous writes are reconciled before they affect future behavior. This distinction mirrors familiar middleware history. Databases did not eliminate the need for data-access middleware; queues did not eliminate messaging middleware; caches did not eliminate consistency and invalidation protocols. Similarly, vector stores and managed memories do not eliminate the need for a layer that specifies how agent experience is transformed into reusable, governed knowledge.

\subsection{The Learning--Governance Gap}

The emerging memory ecosystem exposes a learning--governance gap. Agent-local memories---self-improving agents, coding-agent manifests, skill libraries, and persistent knowledge surfaces~\cite{hermesAgent,cugaAgent,chatlatanagulchai2025manifests,codexAgentsMd,agentsMd,antigravityKnowledge}---sit close to the execution loop. They observe the details of planning, tool use, failures, repairs, and environmental constraints, and therefore can produce operationally useful lessons. However, they tend to be weak on governance: memories are often private to a harness, difficult to inspect, hard to deduplicate, and lacking explicit lifecycle policies.

Platform memories from model and cloud providers~\cite{openaiMemoryFAQ,awsAgentCoreMemory,googleMemoryBank,cloudflareAgentMemory} occupy the opposite position. They are closer to managed infrastructure, with persistence, administrative controls, user-facing memory controls, and integration into broader cloud or model platforms. However, they often emphasize personalization, session continuity, or retrieval over context. They do not necessarily define rich learning semantics for extracting reusable operational guidance from agent trajectories. \rev{The gap is concrete. Suppose a team's coding agent learns, from a painful failure, that one service's staging configuration must never be copied to production. Held in one developer's harness-local memory, the lesson is invisible to teammates' agents, cannot be reviewed or corrected by the team, and vanishes when that developer switches tools; held only in a platform personalization store, it is administrable but was never extracted from the trajectory in the first place. Neither side can both learn the lesson and govern it.}

Memory middleware should bridge these two directions. It should preserve the learning richness of agent-local memory while providing governance hooks for both users and organizations. Users should be able to control what is remembered, forgotten, shared, or scoped to a project. Organizations should be able to publish policies, rules, and guardrails as governed memories that guide agent behavior without being hard-coded into every agent. Multiple agents should be able to draw from a common substrate of preferences, learned experiences, and externally authored guidance while preserving provenance and access control.

\section{A Design Vision for Memory Middleware}
\label{sec:design-space}

\begin{figure}[t]
\centering
\resizebox{\columnwidth}{!}{%
\begin{tikzpicture}[
  font=\sffamily,
  box/.style={draw, rounded corners=2pt, align=center, minimum height=7mm, inner sep=3pt},
  host/.style={box, fill=blue!6, minimum width=24mm},
  store/.style={box, fill=black!6, minimum width=24mm},
  iface/.style={draw, fill=blue!16, rounded corners=2pt, align=center, minimum height=7mm, minimum width=86mm},
  mem/.style={box, fill=green!10, minimum width=24mm, minimum height=6mm},
  arr/.style={<->, >={Latex[length=2.4mm]}, very thick, black!60},
]
\node[host] (h1) at (0,8.6) {Coding\\agent};
\node[host] (h2) at (3,8.6) {Web / GUI\\agent};
\node[host] (h3) at (6,8.6) {Enterprise\\agent};
\node[iface] (top) at (3,7.1) {Agent-facing interface \;\textit{(toward a common standard)}};
\node[draw, very thick, rounded corners=4pt, fill=blue!3, minimum width=92mm, minimum height=28mm] (mw) at (3,4.75) {};
\node[anchor=north, font=\sffamily\bfseries] at ([yshift=-1.2mm]mw.north) {Memory Middleware};
\node[mem] (epi) at (0,5.2) {Episodic};
\node[mem] (sem) at (3,5.2) {Semantic};
\node[mem] (pro) at (6,5.2) {Procedural};
\node[font=\itshape, text width=84mm, align=center] at (3,4.4) {pluggability \,$\cdot$\, interposition \,$\cdot$\, isolation \,$\cdot$\, write-path consistency \,$\cdot$\, sharing + provenance \,$\cdot$\, lifecycle governance};
\node[draw, rounded corners=2pt, fill=orange!12, inner sep=3pt] at (3,3.7) {\textbf{Goals:}\; Accuracy \;$\cdot$\; Cost \;$\cdot$\; Controls};
\node[iface] (bot) at (3,2.4) {Store-facing interface};
\node[store] (s1) at (0,0.9) {Files};
\node[store] (s2) at (3,0.9) {Vector DB\\(pgvector)};
\node[store] (s3) at (6,0.9) {Distributed\\vector DB};
\draw[arr] (h1.south) -- (top.north -| h1);
\draw[arr] (h2.south) -- (top.north -| h2);
\draw[arr] (h3.south) -- (top.north -| h3);
\draw[arr] (top.south) -- (mw.north);
\draw[arr] (mw.south) -- (bot.north);
\draw[arr] (bot.south -| s1) -- (s1.north);
\draw[arr] (bot.south -| s2) -- (s2.north);
\draw[arr] (bot.south -| s3) -- (s3.north);
\end{tikzpicture}%
}
\caption{Memory as a pluggable middleware layer.}
\label{fig:vision}
\vspace{-6mm}
\end{figure}
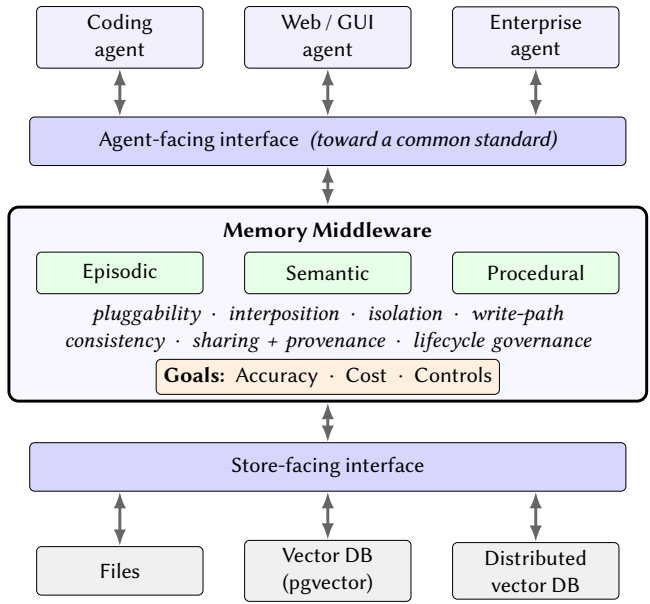

This section develops the vision in three layers: what a memory layer should optimize (its goals), what it must store (the memory types), and the six middleware concerns that operationalize both. Together, they define a research space the middleware community can shape. \rev{The six concerns are not free inventions: they are derived from recurring failure modes of the bespoke systems surveyed in Sections~\ref{sec:problem} and~\ref{sec:related}---context leakage (isolation), memory rot (write-path consistency), silos (pluggability and sharing), and unaccountable influence (provenance and governance)---crossed with the concerns middleware history shows must be extracted for a layer to be operable. We claim this set is sufficient to define the research space, not that it is exhaustive.} Figure~\ref{fig:vision} summarizes the resulting picture.

\subsection{What Should the Goals Be?}

Memory middleware exists to advance three design goals---\emph{accuracy}, \emph{cost}, and \emph{controls}---and these goals are the yardstick against which the rest of this section should be read.

\emph{Accuracy.} Memory should make the agent act better, not merely remember more. This means retrieving the \emph{right} memory for the task, representing it so the signal a query needs is not drowned out (the motivation for purpose-directed gisting), and not degrading behavior with stale, duplicated, or contradictory memory. Accuracy is thus as much about what a memory layer \emph{withholds} as about what it surfaces.

\emph{Cost.} Accuracy has a price, paid in context tokens and execution steps. A memory layer must compress what it stores, recall it lazily rather than dumping it into the prompt, place into the context window only what is needed, and, across deployments, package memory to fit the resource budget of the environment that runs it. Left unmanaged, memory becomes context bloat that is both expensive and, by crowding the window, harmful to accuracy.

\emph{Controls.} To be shared and operated at scale, memory must be governable: isolated by tenant and scope, attributable to its origin, private by default and published only deliberately, and subject to retention, review, and deletion. Controls are what let an organization trust an agent that learns, and what let a user decide what is remembered about them.

Accuracy is the goal the machine-learning community already pursues; \emph{cost} and \emph{controls} are where memory becomes a systems problem, and they are the reason it belongs in middleware rather than inside each agent. The six concerns developed below are the \emph{mechanisms} for advancing these goals, and many serve more than one at once.

\subsection{Support for All Types of Agent Memory}
\label{subsec:memtypes}

A memory layer cannot serve these goals with a single, undifferentiated bag of vectors, because agent experience is not monolithic. It spans \emph{episodic} memory (what happened in a task or session: traces, decisions, errors, outcomes), \emph{semantic} memory (stable facts about the user, project, and environment), and \emph{procedural} memory (how to carry out recurring operations). For conversational agents the substrate must additionally separate conversational turns from task-execution traces, routing the former to semantic memory and the latter to episodic and procedural memory.

Supporting all three in one store is hard because they disagree along nearly every axis a storage layer cares about. They differ in \emph{representation and access} (voluminous time-ordered episodes vs.\ small point-looked-up facts vs.\ task-similarity-retrieved procedures), so no single index suits all three. If commingled in one embedding space, they suffer \emph{retrieval interference}, as high-volume episodic and tool-call content crowds out sparse semantic facts. Their \emph{write semantics} diverge: facts are updated, episodes are immutable, procedures are consolidated. Their \emph{compression} differs: episodes tolerate lossy gists over a durable original, facts must stay precise. Also, they age and are governed differently, demanding distinct isolation and sharing policies. A single episode can even yield \emph{both} a fact and a procedural guideline---a cross-type derivation a flat store cannot express. \rev{The interference failure is equally concrete: in one undifferentiated index, a week of verbose tool-call episodes dominates nearest-neighbor retrieval, so a one-line preference stored months earlier---``always target Java~11''---stops surfacing. It was not forgotten; it was outnumbered.}

\rev{Typing is therefore a structural requirement, not a labeling convention:} a memory middleware must \emph{type} memory and keep the types in separate but linked namespaces, so each is represented, retrieved, reconciled, compressed, and governed on its own terms. This typing reappears inside the concerns below---isolation becomes per-type as well as per-tenant, compression and write-path consistency differ by type, and sharing and governance attach to types differently.

\subsection{Two-Sided Pluggability}

\rev{As illustrated by Figure~\ref{fig:vision}}, the memory layer has two interfaces, not one. \emph{Below}, it must abstract heterogeneous stores---a file store, a relational vector store such as pgvector, a distributed vector database, or a managed memory service. \emph{Above}, it must abstract heterogeneous agent hosts, which differ in how they expose their control loop, tools, prompt context, and traces. The contract must be stable on both sides, so that changing a store touches no agent and supporting a host touches no store. This is the data-access-layer pattern extended with a second pluggable edge facing the agent---what lets learned preferences and repair strategies escape one host's memory format.

What makes this pluggability real is the ingredient the field lacks: a \emph{common interface} that providers implement and agents target. Today every host talks to memory its own way---its own calls to write, recall, and consolidate---though the underlying concepts are nearly identical across agents. Relational databases became interchangeable not because their internals converged but because they all speak SQL; agent memory has no such standard, so each layer stays bespoke and ``pluggable'' remains aspirational. Closing this gap may require more than an API: a new \emph{programming model} for agentic memory---a shared abstraction for how an agent declares what to remember, queries what it knows, and governs it---playing the role the relational model and SQL played for data. Defining that model, and the interface that expresses it, is a central task this Big Idea sets for the middleware community.

\subsection{Host-Native Interposition}

Middleware earns its keep by interposing---intercepting a call, doing work, and getting out of the way. Agent hosts expose different interposition primitives: lifecycle hooks that run before a prompt or after a turn, user-invokable commands, a tool protocol the agent can call, or project-level instruction files. A memory layer must bind to whichever a host provides and present the same logical behavior---recall before the agent acts, capture after---without the agent author writing memory-specific code. The point is not that every host exposes the same mechanism, but that middleware hides these differences behind a stable abstraction, deployable as a sidecar, library, server, tool provider, or plugin while preserving the same operations: retrieve, write, consolidate, publish, delete, inspect.

\subsection{Multi-Tenant Isolation}

Memory is sensitive and contextual: one user's hard-won lesson may be wrong, private, or irrelevant for another. A memory layer therefore needs first-class isolation. Namespaces are a minimal abstraction: memories are scoped by user, team, project, agent, repository, or task, and retrieval is scoped to the active namespace by default. But isolation is load-bearing here, because memories cross sessions and eventually organizations; it must therefore apply not only to storage but to retrieval, consolidation, publication, deletion, and audit. A user should be able to keep a preference private, share it with a project, or delete it; an organization should be able to publish a policy memory to a team without exposing unrelated user memories. A substrate that cannot isolate these scopes will either leak sensitive context or fail to support useful sharing.

\subsection{Write-Path Consistency}

Naive memory is append-only, and append-only memory rots: it accumulates duplicates, stale advice, overgeneralized lessons, and contradictions, then poisons retrieval. A memory layer must treat the write path as a reconciliation problem, deciding whether to add, update, merge, supersede, quarantine, or reject a candidate rather than blindly appending it. This is a consistency and conflict-resolution concern familiar from replicated and versioned systems. For agents, though, the object reconciled is not a data item but a \emph{behavioral influence}: a stale guideline causes future failures, a duplicated preference crowds out better guidance, a contradicted rule makes the agent oscillate. Such guarantees belong in middleware, not in each agent's prompt template.

\subsection{Federated Sharing with Provenance}

The value of memory compounds when it is shared: a guideline one agent learns should be discoverable by others across users, teams, or organizations---but only deliberately. This requires an explicit visibility model (private by default, publishable on purpose) and federation mechanisms by which a namespace subscribes to or synchronizes with other memory sources without copying whole stores or erasing ownership. It also requires provenance: a recalled memory should be traceable to its origin---the trajectory, user, agent, model, or policy that produced it---so that influence on an outcome can be explained and audited. Naming, federation, access control, and provenance are core distributed-middleware concerns; agent memory needs all four. \rev{Sharing also enlarges the security and privacy surface: a shared memory is a channel for poisoning and for leaking personal or proprietary context~\cite{agentpoison2024,memsecsurvey2026}, so admission checks, provenance verification, and default-private visibility are safety mechanisms as much as governance ones.}

\subsection{Lifecycle Governance}

Memories age, getting stale as tools change, policies evolve, APIs are deprecated, or preferences shift. So middleware must support retention, expiration, compaction, auditability, approval, and deletion, which affect the agent's behavior, not just storage. Governance also has multiple authors: memories may be written by the agent, a user, another agent, or an organization publishing rules and guardrails, so the layer must provide policy hooks for who can create, approve, edit, delete, expire, compact, or share each memory. Governance is thus both enterprise-driven (policy and audit controls) and user-driven (control over what is remembered, forgotten, and scoped to a project).

\section{A Reference Architecture and Implementation}
\label{sec:evolve}

We now show that the design space of Section~\ref{sec:design-space} is realizable rather than hypothetical. We describe a reference architecture for memory middleware, and an implementation of it, \evolve~\cite{altkevolveRepo}, that we use in the evaluation. We deliberately organize the description around the \emph{same six concerns} as Section~\ref{sec:design-space}, so that the architecture's \emph{coverage} of the vision is explicit rather than implied.

\begin{figure}[t]
\centering
\resizebox{\columnwidth}{!}{%
\begin{tikzpicture}[
  font=\normalsize\sffamily,
  ringnode/.style={draw, rounded corners=2pt, fill=blue!6, align=center, minimum width=18mm, minimum height=9.5mm, inner sep=2pt},
  agentnode/.style={draw, very thick, rounded corners=2pt, fill=blue!12, align=center, minimum width=18mm, minimum height=9.5mm, inner sep=2pt},
  hub/.style={draw, very thick, rounded corners=2pt, fill=green!8, align=center, minimum width=25mm, minimum height=21mm, inner sep=2pt, font=\normalsize\bfseries\sffamily},
  hostchip/.style={draw, rounded corners=2pt, fill=black!4, align=center, minimum width=23mm, minimum height=7.5mm, inner sep=1.5pt, font=\small\bfseries\sffamily},
  iface/.style={draw, fill=blue!16, rounded corners=2pt, minimum width=70mm, minimum height=7mm},
  ring/.style={-{Latex[length=2mm]}, very thick, black!72},
  plug/.style={-{Latex[length=1.8mm]}, thick, black!55},
  spoke/.style={-{Latex[length=2mm]}, very thick, green!45!black},
  lbl/.style={font=\small\bfseries\itshape\sffamily, text=black!75},
]
\node[hostchip] (hc1) at (-2.85,4.55) {Claude Code};
\node[hostchip] (hc2) at (0,4.55) {Codex};
\node[hostchip] (hc3) at (2.85,4.55) {Claw Code};
\node[iface] (bar) at (0,3.5) {};
\node[lbl] at (bar.center) {agent-facing interface \;(hooks $\cdot$ MCP $\cdot$ skills)};
\node[hub] (hub) at (0,0.45) {Typed Memory\\[-1pt]{\scriptsize\mdseries (store)}\\[2pt]{\footnotesize\mdseries episodic}\\{\footnotesize\mdseries semantic}\\{\footnotesize\mdseries procedural}};
\node[agentnode] (agent)   at (0,2.55)    {Agent};
\node[ringnode]  (capture) at (2.45,1.1)  {Capture};
\node[ringnode]  (distill) at (1.55,-1.45){Extract};
\node[ringnode]  (consol)  at (-1.55,-1.45){Consolidate};
\node[ringnode]  (recall)  at (-2.45,1.1) {Retrieve};
\draw[plug] (hc1.south) -- (hc1.south |- bar.north);
\draw[plug] (hc2.south) -- (bar.north);
\draw[plug] (hc3.south) -- (hc3.south |- bar.north);
\draw[plug] (bar.south) -- (agent.north);
\draw[ring] (agent)   to[bend left=14] (capture);
\draw[ring] (capture) to[bend left=14] (distill);
\draw[ring] (distill) to[bend left=14] (consol);
\draw[ring] (consol)  to[bend left=14] (recall);
\draw[ring] (recall)  to[bend left=14] node[lbl, left=1pt] {inject} (agent);
\draw[spoke] (consol) -- (hub);
\draw[spoke] (hub) -- (recall);
\end{tikzpicture}%
}
\caption{The reference architecture as an experience flywheel: one memory layer that binds to multiple hosts through their native interposition. \rev{Stages are named as in the text: capture, extract, consolidate, store, retrieve, inject.}}
\label{fig:arch}
\end{figure}
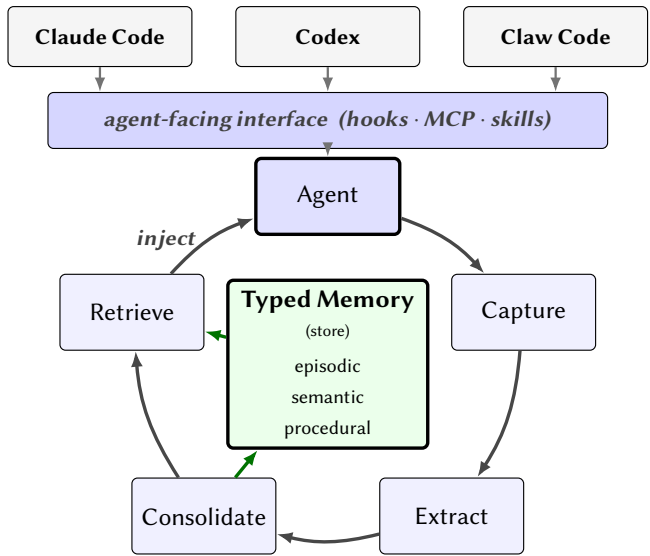

The architecture is a feedback loop---an \emph{experience flywheel} (Figure~\ref{fig:arch})---over five interfaces and one write-path stage. A \emph{capture} interface records agent trajectories; an \emph{extraction} stage distills trajectories into memory objects; a \emph{store} persists them behind a backend abstraction; a \emph{retrieval} interface selects task-relevant memory; and an \emph{injection} path returns that memory to the agent. Between extraction and storage, a \emph{consolidation} stage reconciles new memory with existing memory. \rev{Figure~\ref{fig:arch} shows these stages as one loop---capture, extract, consolidate, store, retrieve, inject---closed through the agent itself.} The architecture prescribes these interfaces and their semantics; it does not prescribe any particular agent, model, or store, and any component may be replaced as long as the interface is honored. The store holds the three types of memory---episodic, semantic, and procedural---in separate namespaces, and represents them compactly as purpose-directed \emph{gists} backed by a durable store of originals (Sections~\ref{subsec:memmodel}--\ref{subsec:gisting}).

\rev{Crucially, the architecture covers the six concerns unevenly, and we say so explicitly: Table~\ref{tab:design-space} records what it realizes today versus what remains open, and its open directions are precisely the research agenda of Section~\ref{sec:agenda}.}

\subsection{A Typed Memory Model}
\label{subsec:memmodel}

The architecture realizes the typed model of Section~\ref{subsec:memtypes}: episodic, semantic, and procedural memory occupy separate namespaces, and the capture interface routes incoming experience to the right type---conversational turns to semantic memory, task-execution traces to episodic and procedural. For example, a trajectory in which an agent repeatedly calls an API with the wrong identifier yields a procedural guideline naming the field the API expects, while a passing remark about a preferred language yields a semantic fact. Separating the types into namespaces keeps them from interfering: high-volume tool activity cannot crowd out user preferences, and procedural recall does not surface incidental chatter. The common unit across types is an \emph{entity}: a typed record carrying content, a trigger or applicability condition, and provenance metadata linking it to its source task and creation mode---a behavioral guideline is one entity type; a user fact and a tool-call summary are others. \rev{Concretely, the persisted record is small: \emph{content} (text or structured data), a \emph{type} tag (guideline, fact, trajectory, policy), and a metadata map carrying the trigger, provenance (source task and session, and whether creation was automatic or manual), ownership and visibility (private by default, public only on publish), and---for guidelines---a category, rationale, implementation steps, and a support count with success/failure evidence that accumulates across consolidations.}

\subsection{\texorpdfstring{\rev{The Agent-Facing Surface and Host Bindings}}{The Agent-Facing Surface and Host Bindings}}
\label{subsec:surface}

\rev{The agent-facing contract is deliberately small: the six logical operations of Section~\ref{sec:design-space}---retrieve, write, consolidate, publish, delete, inspect. In the implementation, the operations an agent invokes mid-task are exposed as tools over a standard tool protocol (MCP, over stdio or SSE): task-conditioned retrieval with type filters and namespace scoping, including a dosage-aware variant that returns an always-on core of high-support guidelines plus the top-$k$ most task-relevant ones; trajectory capture that also triggers extraction (\texttt{save\_trajectory}); fact extraction from conversational turns; manual entity creation; publish and unpublish; and owner-checked delete. Consolidation and inspection are control-plane operations, deliberately kept off the agent's tool surface and exposed instead through a client library, CLI, and web interface---an agent can use memory, but curating memory is a governed, human-visible act.}

\begin{table}[t]
\centering
\footnotesize
\setlength{\tabcolsep}{3.5pt}
\caption{\rev{One memory layer, four host bindings: the native interposition primitive each host exposes, and how the same logical operations attach to it.}}
\label{tab:hosts}
\rev{\begin{tabular}{@{}p{1.5cm}p{3.1cm}p{2.9cm}@{}}
\toprule
Host & Interposition primitive & Operations bound \\
\midrule
Claude Code & Plugin; one-line import in the project instruction file; skills & Recall via the host's native memory, bridged by a mirroring skill; recall audit \\
Codex & Pointer in a global instruction file & Recall-first protocol; saves; recall audit \\
Claw Code & Plugin carrying skills; optional pre-tool-use hook & Recall, learn/save, audit \\
IBM Bob & Global rules file + commands; MCP server in the full deployment & Recall, trajectory capture, entity management via tools \\
\bottomrule
\end{tabular}}
\end{table}

\rev{Table~\ref{tab:hosts} grounds host-native interposition (Section~\ref{sec:design-space}): each host binds the same logical behavior---recall before the agent acts, capture after, with an append-only audit of which memories were consulted and whether they were followed---through whatever primitive that host exposes, and no host's agent code changes. The diversity in the middle column is the point: instruction imports, plugins, skills, hooks, and a tool protocol are all interposition, and the layer absorbs the differences.}

\subsection{From Traces to Reusable Memory}
\label{subsec:extraction}

Raw trajectories are too specific and too verbose to reuse directly. The extraction stage turns them into reusable procedural memory in four steps. (1)~\emph{Task segmentation}: when a trace does not delimit tasks explicitly, the middleware first identifies the distinct tasks the agent actually performed. (2)~\emph{Subtask identification}: each task is decomposed into its constituent subtasks, so knowledge is captured at subtask granularity---far more reusable than the task level, because many unrelated tasks share subtasks such as authenticating with a service or filtering a result set. (3)~\emph{Generalization}: subtask descriptions and their tips are abstracted away from task-specific entities---a named playlist becomes ``a target playlist,'' a specific account becomes ``a user account''---so that memory learned on one task matches structurally similar subtasks elsewhere. (4)~\emph{Clustering and consolidation}: generalized subtasks are clustered by semantic similarity, and the tips within a cluster are merged and deduplicated into consolidated guidance that is more comprehensive than any single instance.

This generalize-then-consolidate discipline is also what makes write-path consistency tractable: reconciliation compares \emph{generalized} candidates against \emph{clustered} prior memory, so duplicates and contradictions are detected at the level of reusable operations rather than verbatim text.

\subsection{Purpose-Directed Gisting and Tiered Storage}
\label{subsec:gisting}

Episodic sessions and tool-call outputs are verbose; storing them verbatim in the retrieval index is costly and dilutes the signal that makes memory findable. The architecture instead represents memory as \emph{gists}---compact, model-produced summaries---under a tiered scheme: the gist is embedded in the vector index, while the full original is retained in a durable store with a pointer from the gist. Because the original is always recoverable, gists can be aggressively compact without risking information loss; the gist is an index, not a replacement.

Critically, gisting is \emph{purpose-directed}: the compression prompt encodes what the gist is for, so the model preserves the signal the downstream query will need rather than defaulting to generic, topic-preserving summarization. A conversational turn bound for semantic memory is gisted to foreground user attributes and preferences; a tool-call interaction bound for procedural memory is gisted to foreground what was retrieved or computed, the parameters used, and the outcome. This matters because a brief preference buried in a long, unrelated message is diluted in a full-passage embedding---purpose-directed gisting concentrates that signal so it dominates retrieval. Tool-call gists are kept in a namespace separate from conversational gists, so the two cannot contaminate each other's retrieval, and each namespace scales independently. At read time the granularity decision is made by the model in context: it inspects the retrieved gist and fetches the full original from the durable store only when the gist is insufficient for the task at hand---adaptive memory fidelity without fixed retrieval rules.

\begin{table}[t]
\centering
\footnotesize
\setlength{\tabcolsep}{3.5pt}
\caption{Coverage of the design space by the reference architecture: realized today vs.\ open.}
\label{tab:design-space}
\begin{tabular}{@{}p{1.9cm}p{2.95cm}p{2.7cm}@{}}
\toprule
Concern & Realized today & Open direction \\
\midrule
Two-sided pluggability & Files, pgvector, Milvus; four hosts & Host/store standards \\
Host-native interposition & Tool protocol, host hooks, plugins & Sidecars; uniform standard \\
Multi-tenant isolation & Per-tenant \& per-type namespaces & Hierarchical, role-based scopes \\
Write-path consistency & Generalize, cluster, consolidate & Formal semantics, transactions \\
Sharing \& provenance & Visibility, git sync, provenance & Federation policy, audit trails \\
Lifecycle governance & Gist compaction; provenance; deletion & Retention, eviction policy \\
\bottomrule
\end{tabular}
\end{table}

\section{Evaluation}
\label{sec:evaluation}

\begin{table}[t]
\centering
\caption{\evolve\ versus a no-memory baseline on AppWorld (Test-Normal split): Task Goal and Scenario Goal completion (\%).}
\label{tab:eval}
\begin{tabular}{lcccc}
\toprule
 & \multicolumn{2}{c}{Baseline (no memory)} & \multicolumn{2}{c}{\evolve} \\
\cmidrule(lr){2-3}\cmidrule(lr){4-5}
Type & Task & Scenario & Task & Scenario \\
\midrule
Aggregate    & 69.6 & 50.0 & 73.2 & 64.3 \\
Easy & 89.5 & 79.0 & 91.2 & 89.5 \\
Medium & 66.7 & 56.2 & 70.8 & 56.2 \\
Hard & 54.0 & 19.1 & 58.7 & 47.6 \\
\bottomrule
\end{tabular}
\end{table}

We evaluate \evolve~\cite{altkevolveRepo} on AppWorld~\cite{appworld2024}, a public multi-step agent benchmark whose tasks require planning, tool use, state tracking, and recovery from errors. Guidelines are extracted from agent trajectories on AppWorld's \emph{train} and \emph{dev} partitions, and both agents are then evaluated on the held-out \emph{test-normal} partition: the baseline runs without \evolve, while the \evolve-enabled agent retrieves the learned guidelines and injects them into each task's context through the memory interface. Both conditions use the same GPT-4.1 agent and reasoning loop, so only the memory layer differs. We observe two signals.

\emph{It works.} Memory improves both of AppWorld's completion metrics, and most where it matters (Table~\ref{tab:eval}). Aggregate \emph{task-goal completion}---per-task success---rises from \textbf{69.6\%} to \textbf{73.2\%}, while the stricter \emph{scenario-goal completion}, which requires \emph{every} task in a scenario to succeed, rises far more, from \textbf{50.0\%} to \textbf{64.3\%} (a 14.3-point gain). The improvement concentrates where tasks are hardest: scenario-goal completion on the \emph{Hard} tier \textbf{more than doubles}, from 19.1\% to 47.6\%. Because the agent's model and reasoning loop are untouched, this gain is attributable to memory supplied as a separable layer, not to a change in the agent itself.

\emph{It is portable.} The same memory layer runs \emph{unmodified} across four independently developed coding-agent hosts\rev{---Claude Code, Codex, Claw Code, and IBM Bob---}binding through each host's native interposition primitive (Section~\ref{sec:evolve}). This exercises the two-sided pluggability and host-native interposition claims in practice: portability across hosts above is demonstrated, not merely asserted, alongside the three swappable storage backends below. We treat these results as evidence of feasibility and signal, not as a complete evaluation; a fuller study is ongoing.

\section{Research Agenda}
\label{sec:agenda}

Treating memory as middleware opens a research agenda the systems community is well positioned to take on. \rev{The agenda has a dependency structure. Two items are foundational: the programming model and common interface of Section~\ref{sec:design-space} (what an agent declares, queries, and governs) and consistency semantics for learned experience, because every other mechanism is specified against them. Admission, forgetting/eviction, deployment-aware packaging, and purpose-directed compression build on those foundations. Isolation and sharing policies, provenance, and portability benchmarks are cross-cutting and can proceed in parallel. We therefore develop the two foundational items in the most depth below.}

\emph{\rev{Foundational:} Consistency models for learned experience.} Traditional storage consistency does not transfer, because memories are generated interpretations of experience rather than ground-truth records. Future work should define consistency models for learned knowledge: when a new memory should replace an old one, when conflicting memories should coexist, and when retrieval should expose uncertainty to the agent. \rev{A concrete starting point is to define, per memory type: a \emph{supersession order} (when a new guideline replaces an old one versus coexisting as an alternative); a \emph{visibility rule} (when a newly written memory becomes retrievable---immediately, after consolidation, or after approval); and an \emph{uncertainty surface} (when retrieval exposes conflicting memories to the agent rather than resolving them silently). Analogues of read-your-writes and monotonic reads exist here---an agent should not lose a lesson it just learned, nor oscillate between contradictory guidance across turns---but the equivalence criterion is behavioral rather than byte-level: two memory states are equivalent when they induce the same agent behavior.}

\emph{\rev{Foundational:} Memory transactions and write policies.} Agent memory writes may involve multiple steps---storing a raw trajectory, extracting candidate memories, checking for conflicts, updating existing memories, and publishing selected items---which require transactional semantics and failure handling; middleware should define atomicity, rollback, and approval policies for memory writes. \rev{The natural transaction unit is the staged write pipeline of Section~\ref{sec:evolve}---capture, extract, consolidate, publish---with defined rollback points: a failed consolidation must not leave half-merged guidance retrievable, and an unapproved candidate must not influence behavior while it awaits review.}

\emph{\rev{Building on the foundations:} Memory admission and quality-weighted learning.} Not every experience deserves to become memory: an agent that learns indiscriminately internalizes the mistakes of failed runs as readily as the insights of successful ones. Middleware therefore needs an \emph{admission} stage that scores a candidate experience before distilling it and weights the result by source quality, combining heterogeneous, partly reliable signals---task outcomes, agent self-reflection, failure counts, and, most valuably, \emph{successful recovery from failure}, which encodes a reusable strategy a clean success never exercises. Open problems include calibrating these signals and re-scoring memories as their downstream reuse is observed.

\emph{\rev{Building on the foundations:} Forgetting, eviction, and compaction.} Long-running agents will accumulate more experience than can be retrieved or governed manually, so middleware must decide what to retain, what to compact, and what to forget. Unlike cache eviction, memory eviction changes agent behavior, and policies should therefore consider recency, utility, provenance, confidence, sensitivity, and staleness.

\emph{\rev{Building on the foundations:} Deployment-aware packaging and placement.} Agents run across deployment tiers---VMs with gigabytes of RAM, containers with hundred-megabyte limits, and serverless functions with tight cold-start budgets---yet memory is typically one remote service, oblivious to where the agent runs. Treating memory as middleware invites a packaging layer that emits the \emph{same} memory in tier-appropriate formats, raising systems questions new to agent memory: compression-aware indexing via embedding quantization, automatic format selection under resource constraints, distribution through init containers and sidecars, and \emph{budget-adaptive tiered retrieval} that pages in from a remote tier only when local quality is low and the budget permits.

\emph{\rev{Building on the foundations:} Purpose-directed compression and adaptive fidelity.} The reference architecture compresses memory with purpose-directed gists (Section~\ref{subsec:gisting}); making this principled raises open questions: how to \emph{infer} a memory's downstream purpose at capture rather than hand-writing a prompt per type, whether one memory can serve multiple purposes, how to bound information loss for high-stakes memories, and what \emph{adaptive-fidelity} policy decides at read time whether a gist suffices or the full original must be fetched.

\emph{\rev{Cross-cutting:} Isolation and sharing policies.} Namespaces are a starting point, but enterprise deployments need richer policy models: hierarchical scopes, role-based access, task-specific sharing, team-level memory, and promotion workflows that allow useful private memories to become shared organizational knowledge.

\emph{\rev{Cross-cutting:} Provenance and auditability.} Generated memories influence future decisions, so systems must answer where a memory came from, which trajectory and model produced it, who approved it, and which behaviors it later influenced---and preserve that lineage even when memories are compacted or merged.

\emph{\rev{Cross-cutting:} Portability benchmarks.} Current evaluations often couple a memory system to a particular agent, benchmark, and backend. The middleware community should develop portability benchmarks that evaluate memory systems independently of any single agent framework---testing whether memory can move across agents, stores, protocols, and deployments while preserving behavior and governance.

\section{Related Work}
\label{sec:related}

Agent memory is a fast-moving area; we group the most relevant work by where it sits relative to our argument. Prior work contributes memory \emph{algorithms}, \emph{architectures}, or \emph{products}; we argue for the middleware \emph{boundary} that lets them interoperate.

\emph{Experiential and procedural memory.} A large body of work makes agents improve from their own execution without weight updates: reflection on failures~\cite{shinn2023reflexion}, simulated experience and episodic recall~\cite{park2023generativeagents}, skill induction~\cite{wang2023voyager}, natural-language insight extraction~\cite{expel2023}, induced task workflows~\cite{awm2024}, test-time strategy memories~\cite{cheatsheet2025}, evolving context playbooks~\cite{ace2025}, and trajectory-distilled reasoning or procedural memory~\cite{reasoningbank2025,reme2025,memp2025,memento2025,memoryr1,fang2026trajectory}. These show experience helps; our contribution is the substrate that lets any such method plug into many hosts and stores instead of living inside one agent.

\emph{Memory architectures and typed, compressed memory.} Others propose memory \emph{representations}: OS-style paging of context~\cite{packer2023memgpt}, test-time neural long-term memory~\cite{titans2025}, episodic memory control~\cite{larimar2024}, hierarchical and agentic memory graphs~\cite{hmem2025,xu2025amem,mirix2025}, gist compression of very long contexts~\cite{readagent2024}, and typed memory representations~\cite{memir2026,memanto2026,dimmem2026}. We adopt typed, separated memory and purpose-directed gisting but treat them as middleware concerns---contracts and namespaces---not a single model; whether agent memory is a database is itself debated~\cite{orogat2026agentmemorydatabase}.

\emph{Write-path, admission, and retrieval policy.} Recent systems ask \emph{what} to store and \emph{when}: embedding-based write admission~\cite{memrouter2026}, recurrence-triggered consolidation~\cite{recmem2026}, causal rather than similarity-based selection~\cite{causalmem2026}, closed-loop tuning of memory operations~\cite{evolvemem2026}, and empirical studies of memory management~\cite{memmgmt2025}. These motivate our write-path-consistency and admission agenda, framed as policies a memory layer should own rather than each agent.

\emph{Trust, provenance, and portability.} Persistent memory introduces new attack surfaces and governance needs: memory poisoning~\cite{agentpoison2024}, lineage and post-hoc audit~\cite{memlineage2026,memaudit2026}, governed evolving memory~\cite{governingmem2026}, security surveys~\cite{memsecsurvey2026}, evidence-tiered provenance~\cite{tiermem2026}, and provenance-verified transfer across heterogeneous agents~\cite{portableagentmemory2026}---aligning with our \emph{controls} goal and our isolation, provenance, and portability agenda.

\emph{Skills as procedural memory.} Procedural memory increasingly materializes as reusable skills discovered, refined, and curated from experience~\cite{wang2023voyager,skillweaver2025,mementoskills2026,skillos2026}, though skills can also add overhead or harm~\cite{whenskills2026}; our procedural memory and consolidation are complementary, and governed skill promotion is part of our agenda.

\emph{Memory services and platforms.} Productized memory layers~\cite{mem0}, managed platform memories~\cite{claudeDreams,copilotMemory,awsAgentCoreMemory,googleMemoryBank,cloudflareAgentMemory,oracleAgentMemory}, framework primitives~\cite{langmem}, and file-native memories~\cite{agentsMd,codexAgentsMd,openclawSkills} are strong on storage and governance but are typically bound to one host or cloud, with semantics centered on personalization and session continuity. We instead offer a host- and store-portable substrate with the cross-cutting guarantees---two-sided pluggability, isolation, consistency, sharing, and lifecycle governance---none of these expose uniformly.

\section{Conclusion}

Agents are becoming persistent software actors, but their memory remains fragmented application logic. We have argued that it should instead be a first-class middleware substrate---one that interposes between agents and stores, learns from experience, and governs the result. \evolve~\cite{altkevolveRepo} shows that an initial subset is practical and improves agent reliability; the broader challenge---defining the abstractions, consistency models, policies, and benchmarks that make memory a reusable foundation for self-improving agents---is now open to the systems community.

\balance
\bibliographystyle{ACM-Reference-Format}
\bibliography{references}

\end{document}